\documentclass{article}
\usepackage{ijcai22}

\usepackage{times}
\usepackage{soul}
\usepackage{url}
\usepackage[hidelinks]{hyperref}
\usepackage[utf8]{inputenc}
\usepackage[small]{caption}
\usepackage{graphicx}
\usepackage{amsmath}
\usepackage{amsthm}
\usepackage{booktabs}
\usepackage{algorithm}
\usepackage{algorithmic}
\usepackage{multirow}

\title{Automatic Block-wise Pruning with Auxiliary Gating Structures for Deep Convolutional Neural Networks}

\author{
Zhaofeng Si$^1$
\and
Honggang Qi$^1$\and
Xiaoyu Song$^{2}$
\affiliations
$^1$University of Chinese Academy of Sciences\\
$^2$Portland State University
\emails
sizhaofeng19@mails.ucas.ac.cn,
hgqi@ucas.ac.cn,
songx@pdx.edu
}

\begin{document}

\maketitle

\begin{abstract}
  Convolutional neural networks are prevailing in deep learning tasks. However, they suffer from massive cost issues when working on mobile devices. Network pruning is an effective method of model compression to handle such problems. This paper presents a novel structured network pruning method with auxiliary gating structures which assigns importance marks to blocks in backbone network as a criterion when pruning. Block-wise pruning is then realized by proposed voting strategy, which is different from prevailing methods who prune a model in small granularity like channel-wise. We further develop a three-stage training scheduling for the proposed architecture incorporating knowledge distillation for better performance. Our experiments demonstrate that our method can achieve state-of-the-arts compression performance for the classification tasks. In addition, our approach can integrate synergistically with other pruning methods by providing pretrained models, thus achieving a better performance than the unpruned model with over 93\% FLOPs reduced.
\end{abstract}

\section{Introduction}

Convolutional neural networks (CNNs) have achieved amazing performance in different domains including image classification,
object detection 
and semantic segmentation. 
One of the most important supportive factors lies on the complexity of the network structure, which allows the models to learn sufficient knowledge within certain datasets to generalize well on real-world scenes. While gaining promotion in performance, however, such complex structures can also suffer from high computation cost and storage cost in certain applications such as applications on mobile devices and other platforms with constrained resources. To tackle this problem, researchers have developed a series of methods called model compression to cut down the cost while using CNNs.

\begin{figure}[t]
	\centering
	\includegraphics[width=0.9\columnwidth]{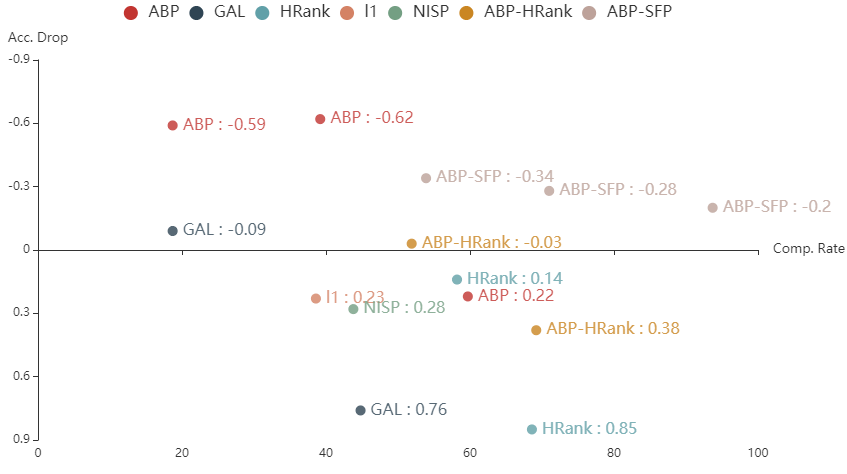} 
	\caption{Visualization of pruning results of our ABP method and other state-of-the-arts methods. The y-axis is reversed so that points in upper position means high performance and minus value in accuracy drop means increase in accuracy compared to baseline model.}
	\label{fig0}
\end{figure}

Dominating methods in model compression can be mainly divided into four types: network pruning \cite{2015Deep}, weight quantization \cite{DBLP:conf/cvpr/TungM18}, decomposing \cite{2014Speeding} and knowledge distillation \cite{2015Distilling}, among which network pruning is a traditional yet effective way of model compression.

As the depth of neural works going deep, demands on light-weighted models are becoming much more severe, Motivating more researches on pruning methods. Typical pruning methods can be divided into structural pruning \cite{2015Structured} and non-structural pruning \cite{2015Deep} according to the granularity of pruning. Non-structural pruning tries to zero-out the unimportant parameters within layers based on a ranking algorithm, while structural pruning cuts off entire structure (layer, convolutional kernel, channel etc.) iteratively or simultaneously. Non-structural pruning is advantageous on reducing storage cost, but cannot release computational cost effectively since the computation would still exist even though some of the weights in a layer is zeroed out, while structural pruning can avoid that problem by removing the whole structure, but is would suffer from greater loss in performance than non-structural pruning. In this work we focus on structural pruning with a large granularity, i.e. block-wise pruning, in an iterative manner. Block here refers to basic units the CNN is formed with, for example, residual blocks in ResNet series and single convolutional layers in VGG style CNN.


\begin{figure*}[t]
	\centering
	\includegraphics[width=0.8\textwidth]{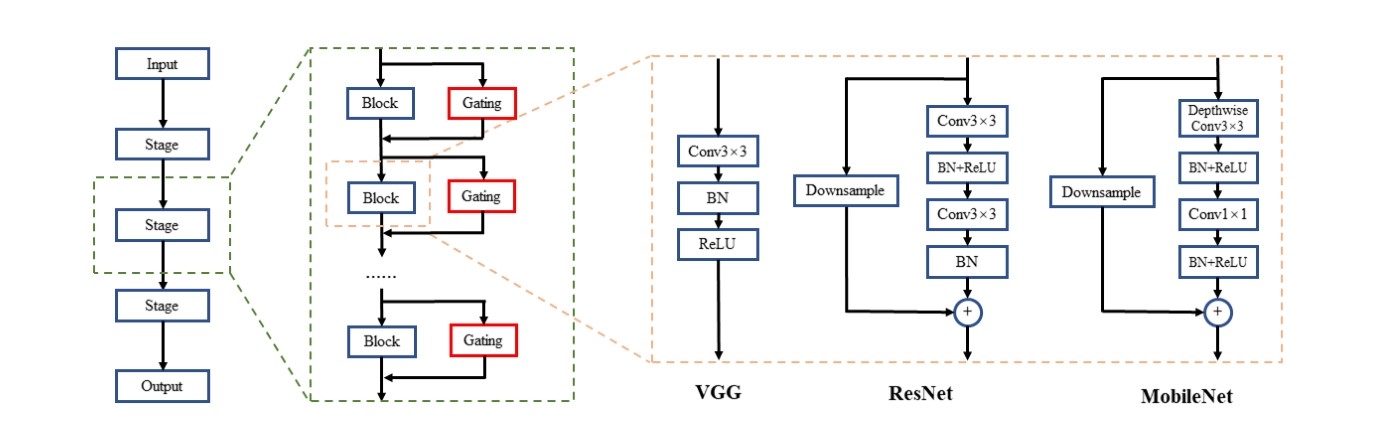} 
	\caption{Overall architecture of our method. We illustrate the definition of block in our work with different networks. Gating modules are marked red which are key component of the pruning algorithm and is illustrated in detail in Figure 3.}
	\label{fig1}
\end{figure*}

SkipNet \cite{DBLP:conf/eccv/WangYDDG18} is a popular method of dynamic neural network, where a control module is proposed to decide blocks to skip during inference according to different instances. However, this method can only cut down computing cost during inference by skipping some of the blocks without removing them, which means the issue of storage cost is even worse with the extra cost brought by control modules. We show that dynamic neural network methods can be incorporated into network pruning in voting style to decide the least important blocks and remove them completely. The overall architecture is shown in Figure 2. 

We then designed a training schedule for proposed structure, which consists of three stages including warm-up, iterative pruning and fine-tuning. In the warm-up stage, the model is trained for a few epochs to grab knowledge from the dataset. Then in the pruning stage, blocks are iteratively pruned according to marks derived from each block with an auxiliary structure, with several epochs of training to restore performance after pruning in each iteration. It is worth mentioning that knowledge distillation \cite{2015Distilling} is used in this stage in self-distillation manner to further enhance the performance, where models from previous iteration is used as teacher model in the following iteration. After pruning stage, the model is fine-tuned with a learning rate schedule to obtain the final pruned model.

As a pruning method with large granularity, our method can reduce storage cost and computational cost effectively in block-wise, but redundancy inside blocks is still intact. We show that our pruned model can be further used in other pruning methods with small granularity as a pre-trained model by conducting experiments that result in surprisingly good performance, which is an important contribution of our method.

The contributions of our work are summarized as follow:
\begin{itemize}
\item We proposed a network pruning algorithm in block-wise granularity (Automatic Block-wise Pruning, ABP) which prunes a CNN using gating structure based on the idea of dynamic neural network.
\item We designed a three-stage training schedule (warming-up, pruning and fine-tuning) incorporating knowledge distillation for proposed structure for iterative pruning.
\item We demonstrate that our method can provide pretrained model with  prior knowledge for other pruning methods. Results show that this combination can achieve surprising compression rate of 93.68\% with even higher performance than unpruned model. 
\end{itemize}

\section{Related Works}
\subsection{Network pruning}
Network pruning is a traditional and straightforward method for model compression, which operate directly on weights and architecture of the network to cut down storage cost and computational cost. Pruning methods can be divided in to structured pruning and unstructured pruning, where structured pruning tries to prune entire structure like convolution kernel or layer, which is the main focus in our work. Anwer et.al \cite{2015Structured} proposed a schema of structured pruning in all granularity, and involved particle filter into pruning schedule. AutoCompress \cite{DBLP:journals/corr/abs-1907-03141} developed an automatic structured pruning schedule that can compress at extremely large compress rate. Recently several works \cite{DBLP:conf/aaai/RoC21} have focused on large granularity of pruning like layer-wise, which is closely related to our work.

\subsection{Dynamic neural network}
Dynamic neural network has been rapidly developed in recent years, where parameters or architecture of the network are adapted to different inputs during inference \cite{DBLP:journals/corr/abs-2102-04906} to reach a balance between model performance and efficiency. Instance-wise dynamic neural network methods \cite{DBLP:conf/aaai/LiuD18} are the most common methods where the network would change in architecture and parameters according to the input instance. D$^2$NN \cite{DBLP:conf/aaai/LiuD18} proposed to select the execution path of DNN with several control nodes, ending up with different topological structures for different instances, while SkipNet \cite{DBLP:conf/eccv/WangYDDG18} realized dynamicity without changing the entire topology of the architecture by imposing control module to all blocks to decide whether to skip the block. Our work was inspired by the "skipping" strategy, based on which we extended the idea of dynamic neural network to network pruning with a voting-like method.

\subsection{Knowledge distillation}
Knowledge distillation was firstly introduced by Hinton \cite{2015Distilling} who then opened up a new research direction of model compression. This kind of methods aims at training a light-weighted model using the knowledge in the form of soft label \cite{2015Distilling} or intermediate representation \cite{DBLP:conf/aaai/JiHP21} from a cumbersome network with teacher-student structure. Aside from the usage in model compression, knowledge distillation methods are also used in enhancing performance in various tasks \cite{DBLP:journals/corr/abs-2002-10345}. Researchers have also developed other forms of knowledge distillation beyond teacher-student structure including self-distillation \cite{DBLP:journals/corr/abs-2002-10345}, multi-teacher distillation \cite{DBLP:journals/corr/abs-1903-04197} and mutual learning \cite{DBLP:conf/cvpr/ZhangXHL18}, where self-distillation would benefit the iterative training schedule proposed in our work.

\section{Automatic Block-wise Pruning}
\subsection{Problem formulation}

\begin{figure}[t]
	\centering
	\includegraphics[width=0.9\columnwidth]{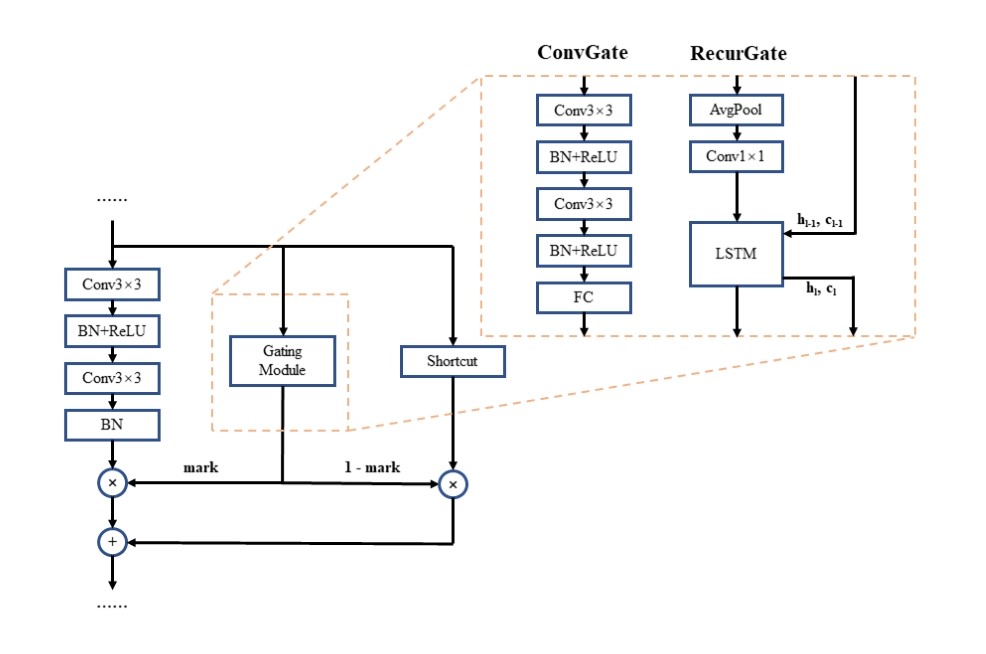} 
	\caption{Gating layers in our architecture.}
	\label{fig2}
\end{figure}

Given a convolutional neural network, our purpose is to prune the network for efficient usage. Consider an $N$-block CNN model, its blocks can be represented as ${B}=\left \{ {B}_{1},{B}_{2},...,{B}_{N} \right \}$, and their inputs and outputs are $\mathcal{I}=\left \{ \mathcal{I}_{1},\mathcal{I}_{2},...,\mathcal{I}_{N} \right \}$ and  $\mathcal{O}=\left \{ \mathcal{O}_{1},\mathcal{O}_{2},...,\mathcal{O}_{N} \right \}$ respectively. Forward pass through a block can be represented as:

\begin{equation}
\mathcal{I}_{l+1}=\mathcal{O}_{l}={B}_{l}\left ( \mathcal{I}_{l} \right )\quad for\quad l = 1,2,...,N
\end{equation}

Denoting the input of the network as $x$, we have the forward pass of the network before classifier as:

\begin{equation}
    \begin{split}
        f={B}_{N}\left (...{B}_{2}\left ({B}_{1}\left ( x \right ) \right) \right)=concat(B_l) \\ for\quad l = 1,2,...,N\label{2}   
    \end{split}
\end{equation}
 where $concat(B_l)$ means concatenating these blocks by taking the output of former block as the input of current block, and $f$ is the final feature map before classifier.

In this paper we prune the network in a large granularity, i.e. block-wise. To achieve block-wise pruning, we calculate a criterion for each block as its importance mark, which is determined by gating block in this work, and decide the unpruned block set $\Omega$ accordingly. The target network is formulated as:

\begin{equation}
f=concat(B_l)\quad for\ \ B_l \ \ in \ \ \Omega
\end{equation}

\subsection{Network architecture}

In this section we will introduce in detail about our proposed Automatic Block-wise Pruning method. The term "block" used in this paper refers to the basic structure that forms the entire network, for example, residual block in ResNet \cite{2016Deep} series and convolution layer in VGG \cite{2014Very} style networks. In this section, we will take ResNet architecture as an example.

Inspired by SkipNet \cite{DBLP:conf/eccv/WangYDDG18}, we impose each block ${B}_{l}$ with a gating module ${G}_{l}$, which is shown in Figure 3. Gating module takes the input of current block, and outputs a scalar value for the corresponding block as an importance mark, which can be represented as $m_{l}={G}_{l}(\mathcal{O}_{l-1})$. The mark would then be used as a weighting value for current block to generate the input of the next block $B_{l+1}$. After adding the gating module, (1) can be revised to:

\begin{equation}
	\begin{split}
		\mathcal{I}_{l+1}=\mathcal{O}_{l}\times m_{l} + \mathcal{I}_{l}\times (1-m_{l})  ={B}_{l}\left ( \mathcal{I}_{l} \right )\times {G}_{l}(\mathcal{I}_{l}) \\+ \mathcal{I}_{l}\times (1-{G}_{l}(\mathcal{I}_{l}))\quad  for\quad l = 1,2,...,N
	\end{split}		
\end{equation}

The design of gating module still follows SkipNet \cite{DBLP:conf/eccv/WangYDDG18}, where two types of gating modules are used to calculate importance mark for current block, namely ConvGate and RecurGate. ConvGate consists of two convolution layers to extract feature and a fully connect layer to calculate mark, which is shown in Figure 3. RecurGate is more light-weighted, with one convolution layer attached to each block and an LSTM for the whole network, where LSTM takes the feature extracted from each block by convolution layer as input, and then output the importance mark accordingly. Since gating modules are removed from the architecture after training stage, we don't have to concern about the cost brought by them like SkipNet. Different from SkipNet where the outputs of gating modules are rounded to 0 or 1 in SkipNet to reduce computation, which can cause non-differentiable problem in pruning, we directly use the outputs of gating modules as a metric for pruning.

\subsection{Training schedule}

Together with the ABP architecture, we propose a three-stage training schedule including warm-up, iterative pruning and fine-tuning to obtain a pruned model, which is shown in Figure 4.
In the warm-up stage, ABP model is trained for a few epochs to match the ground-truth label. Denoting the ground-truth label as $y$, the loss function used in this stage is formulated as:

\begin{equation}
\mathcal{L}_{stage1}=\mathcal{L}_{CE}\left(s\left( l \right), y\right)
\end{equation}
where $l$ is the logits output by ABP model, and $s\left( l \right)$ is softmax function taking logits as input.

After warming-up, pruning is implemented in an iterative manner. Concretely, in each iteration, a final mark is calculated by averaging marks generated by gating module for all instances. Assuming there are $S$ training instances, the final mark $\mathcal{M}_{l}$ for block $B_l$ is calculated as:

\begin{equation}
\mathcal{M}_{l}=\sum_{i}^{S}m_{l}^{i}
\end{equation}

\begin{figure*}[t]
	\centering
	\includegraphics[width=0.8\textwidth]{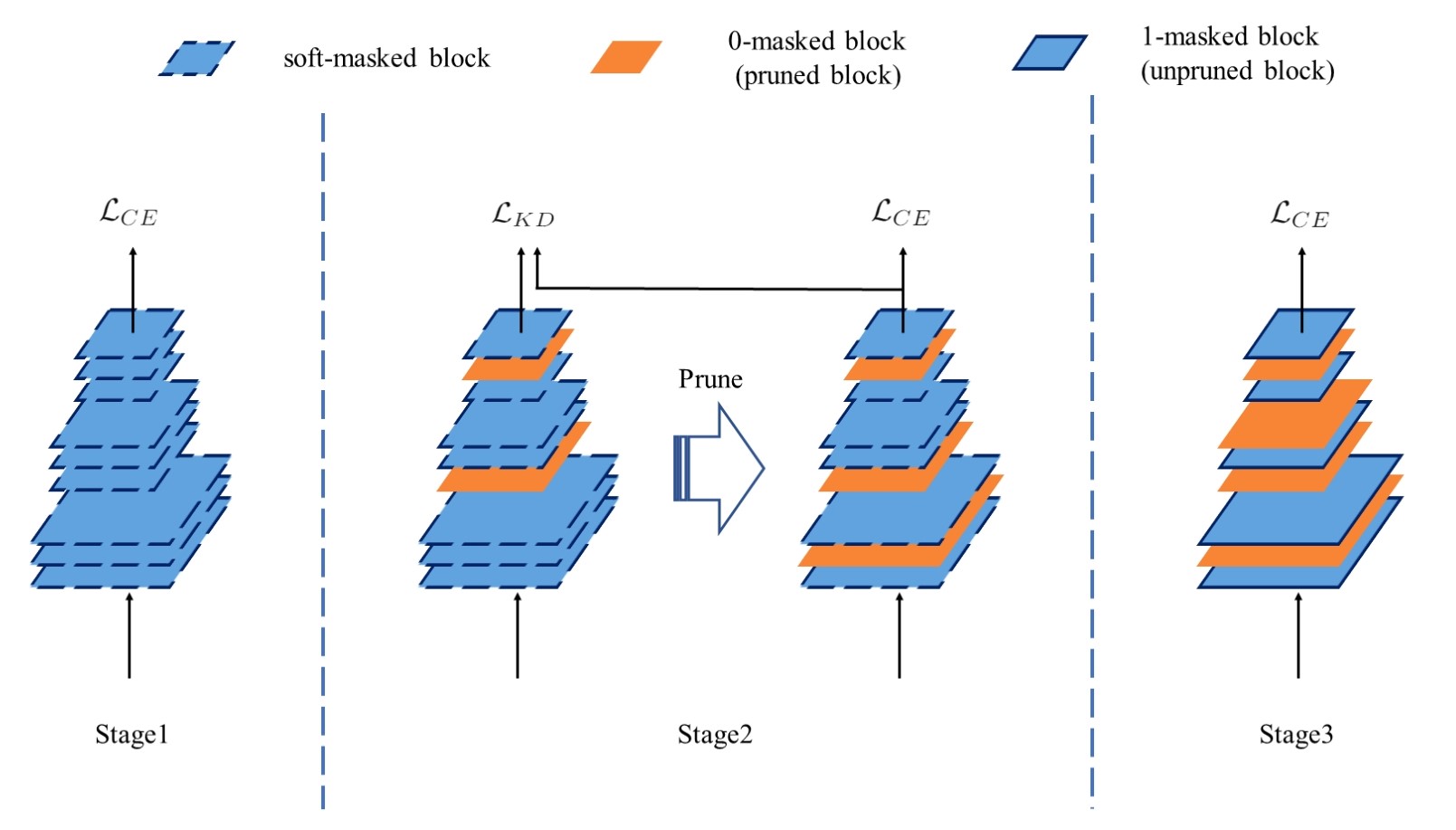} 
	\caption{Example of the whole training schedule. Block with dashed line means the output of corresponding gating module is a floating point number in [0, 1]. In stage 2, the pruned block is skipped by setting its mask to 0, and in stage 3, the masks of remaining blocks are fixed to 1 which is represented by block with solid line.}
	\label{fig3}
\end{figure*}

By calculating the final marks, $k$ blocks with lowest marks can be pruned by applying a zero mask on the output of corresponding blocks, where $k$ is a hyper-parameter defined according to model size. We call it voting-like strategy because we calculate the final mark by summing across all samples, where larger mark represents larger contribution of the block to the task. After pruning, we train the pruned model for a few epochs to restore performance for the next iteration.

We also incorporate knowledge distillation \cite{2015Distilling} into training in this stage to enhance training performance. In our setting model whose from previous iteration capacity is larger is used as teacher model, providing soft target for current iteration to grab latent information while training, which is shown in Figure 3. This iterative training pattern can improve the performance of training while avoiding efficiency decay of distillation brought by large capacity gap. The target of knowledge distillation can be formulated as:

\begin{equation}
\mathcal{L}_{KD}=\mathcal{L}_{KL}\left ( s(l^i;\tau ), s(l^{i-1};\tau ) \right )
\end{equation}
where $\mathcal{L}_{KL}$ is KL divergence, $i$ is the index of current iteration and $\tau$ is temperature term to soften the label for learning efficiency. Then the total loss used in this stage is:

\begin{equation}
    \mathcal{L}_{stage2}=\mathcal{L}_{CE} + \lambda \mathcal{L}_{KD}
\end{equation}
where $\lambda$ is a balance factor, which is set to $\tau^2$ in this paper to balance the magnitude of two loss terms. A pruned model would be obtained by setting the masks of the rest unpruned blocks to 1, and then we propose to retrieve the performance by adding a fine-tuning stage with a learning rate schedule. The entire training process is shown in Algorithm 1 in Supplementary Material.

After training we now have a well-trained network and a gating mask $\mathcal{S}$, where 0 mask means that the corresponding block is pruned, while 1 mark stands for unpruned block . Post-processing procedure is needed before further usage by removing gating modules and pruned blocks in architecture and in parameters. By doing so, we can reduce both computational cost and storage cost when using the model in real-world scenes.

\section{Experiments}
\subsection{Experiment settings}
To validate our proposed method, we implemented ResNet \cite{2016Deep} series models specially for small images ($32 \times 32$) in CIFAR datasets. We set the hyper-parameters according to different architecture and pruning ratio, ensuring the training procedure of Stage2 would contain 2 or 3 iterations. During iterative pruning stage, we would train each model after pruning for 30 epochs, with learning rate decayed at 20th epoch, which is then reset on entering next iteration. For other hyper-parameters, we set initial learning rate $\alpha$ to 0.1 and make it decay on entering Stage2 and Stage3, and set temperature term $\tau$ to 3. All models are trained on a single RTX2080 GPU with batch size as 128 using an SGD optimizer. In our experiments we compare with other state-of-the-art pruning methods including SFP \cite{ijcai2018-309}, l1 \cite{DBLP:journals/corr/LiKDSG16}, GAL \cite{DBLP:conf/cvpr/LinJYZCYHD19}, AMC \cite{DBLP:conf/eccv/HeLLWLH18}, FPGM \cite{8953212}, NISP \cite{DBLP:conf/cvpr/Yu00LMHGLD18}, HRank \cite{DBLP:conf/cvpr/LinJWZZ0020}.

\subsection{Result on CIFAR10}
We compress networks of different depth using proposed method on CIFAR-10 dataset, and the results are shown in Table 1. The results of SOTA methods listed here come from the original papers, and we arrange them according to the drop in FLOPs after pruning, followed with the accuracy drop using each method. As shown in Table 1, our method can achieve better performance than those with similar compression rate, for example, when using ResNet32, ABP-0.6 provides smaller accuracy drop (0.37\% vs 0.70\%) with higher compression rate compared to SFP (61.7\% vs 53.2\%); while in ResNet56 experiments, ABP-0.2 can increase the performance of the network to a large extent (0.91\%) with 15\% of the FLOPs pruned, and ABP-0.6 achieves much better performance in accuracy than GAL (0.08\% vs 1.68\% drop in accuracy) with the same compression rate. As for large scale networks (ResNet110), ABP does better in removing redundancy (0.59\% increase in accuracy) when pruning with a small pruning rate, and brings less accuracy drop (0.02\% vs 0.76\%) with a higher pruning ratio compared to GAL (59.7\% vs 44.8\%).

\begin{table}[t]
    \centering
    \begin{tabular}{c|lrr}
    \toprule
    Arch. & Methods & FLOPs↓(\%) & Acc↓(\%) \\ 
    \midrule
    \multirow{5}{*}{RN32} & \textbf{ABP-0.4} & \textbf{41.1}  & \textbf{-0.33} \\
    & SFP     & 41.9           & 0.55              \\
    & FPGM    & 53.2           & 0.70               \\
    & \textbf{ABP-0.6}    & \textbf{61.7}           & \textbf{0.37}      \\ 
    & \textbf{ABP-SFP}    & \textbf{80.8}           & \textbf{0.97}      \\ 
    \midrule
    \multirow{12}{*}{RN56} & \textbf{ABP-0.2} & \textbf{15.0}  & \textbf{-0.91}    \\
    & l1      & 27.6           & -0.04             \\
    & GAL-0.6 & 37.6           & -0.12             \\
    & \textbf{ABP-0.4} & \textbf{37.6}  & \textbf{-0.34} \\
    & NISP    & 43.6           & 0.03              \\   
    & AMC     & 50.0             & 0.90               \\
    & HRank   & 50.0             & 0.09              \\
    & GAL-0.8 & 60.2           & 1.68              \\
    & \textbf{ABP-0.6} & \textbf{60.2}  & \textbf{0.08}     \\
    & HRank   & 74.1           & 2.54              \\ 
    & \textbf{ABP-HRank} & \textbf{79.2}  & \textbf{2.13}     \\
    & \textbf{ABP-SFP} & \textbf{79.6}  & \textbf{-0.18}     \\
    \midrule
    \multirow{12}{*}{RN110}& GAL-0.1 & 18.7           & -0.09             \\
    & \textbf{ABP-0.2} & \textbf{18.7}  & \textbf{-0.59}    \\
    & SFP     & 28.2           & -0.25             \\
    & l1      & 38.6           & 0.23              \\
    & \textbf{ABP-0.4} & \textbf{39.2}  & \textbf{-0.62}    \\
    & NISP    & 43.8           & 0.28              \\
    & GAL-0.5 & 44.8           & 0.76              \\
    & \textbf{ABP-0.6} & \textbf{59.7}  & \textbf{-0.06}     \\
    & HRank   & 68.6           & 0.85              \\ 
    & \textbf{ABP-HRank} & \textbf{69.2}  & \textbf{0.38}     \\
    & \textbf{ABP-SFP} & \textbf{93.7}  & \textbf{-0.20}     \\
    \bottomrule
    \end{tabular}
\caption{Pruning results compared to SOTA methods. All data are derived from original papers. The number after method name (e.g. 0.4 in ABP-0.4) represents pruning ratio, and minus values in Accuracy drop columns means increase in accuracy after pruning. Method used in this paper is marked bold. ABP-SFP and ABP-HRank are experiments where our methods are incorporated with other SOTA methods, which is demonstrated in detail in Supplementary Material.}
\label{table1}
\end{table}

\begin{figure}[t]
	\centering
	\includegraphics[width=0.99\columnwidth]{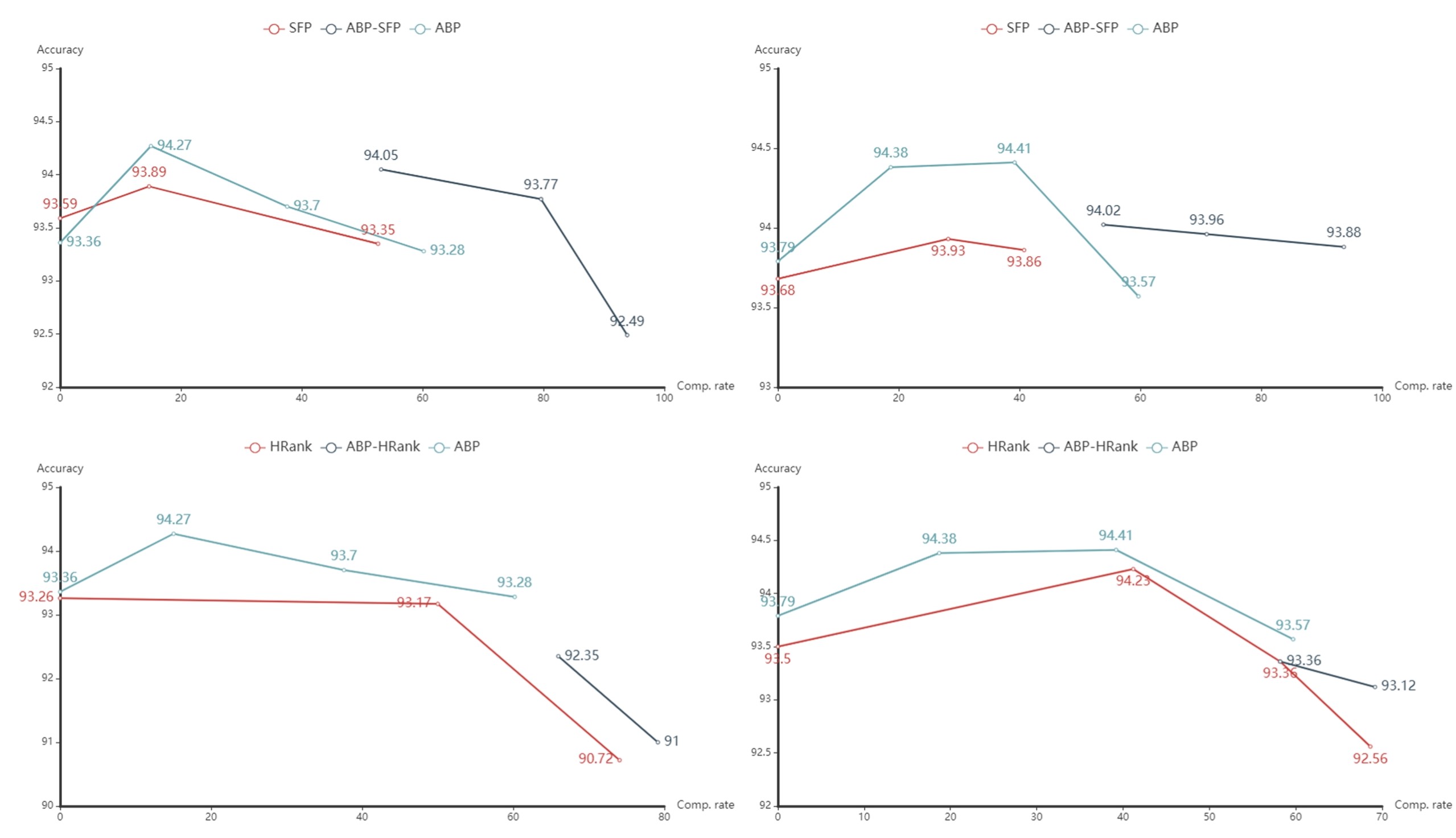} 
	\caption{Performance comparison when incorporating with other SOTA methods including SFP (top left and top right) and HRank (bottom left and bottom right) by providing them with pretrained models. This figure involves ResNet56 (left side) and ResNet110 (right side).}
	\label{fig5}
\end{figure}

Since ABP is a pruning method with very large granularity, it can be easily incorporated with other pruning methods with smaller granularity like channel-wise pruning. We also list some of the results when using our pruned model as pre-trained model (denoted as ABP-SFP and ABP-HRank). Only models with extremely high compression rate are listed in Table 1 for comparison. When incorporating with our method, SFP constantly achieves very high compression rate in all settings, with even higher performance than baseline models on ResNet56 and ResNet110. ABP-HRank failed to achieve comparable performance with ABP-SFP though, it outperforms other SOTA methods with similar compression rate.

For the rest of the results, an illustration is provided in Figure 5, where red lines indicate the original SFP and HRank methods while green lines are original ABP methods, and black lines represent incorporated models. It is obvious that green lines stay constantly above red lines, indicating that ABP method outperform these methods in accuracy under different compression rates. In the case of black lines, they are completely on the right side of red lines, showing that incorporated methods can achieve better compression rate with similar accuracy. 

Results above have shown that as a large granularity pruning method, ABP can provide well trained pretrained model to other pruning methods for further pruning while achieving comparable performance with SOTA methods alone. Detailed experiments on both CIFAR10 and CIFAR100 datasets are provided in Supplementary Material, as well as further investigation on the influence of pruning rate and the type of gating structure.




\subsection{Importance marks}
In order to verify the effectiveness and consistency of our proposed method, we visualize importance marks after stage1 as shown in Figure 6. Every mask is divided into 3 rows, which indicate three different stages in ResNet. As can be seen in Figure 4, top blocks tend to have larger marks while the marks of middle blocks tend to be smaller, which is common in most cases. This phenomenon can be interpreted as deeper layers tend to extract discriminative features, which is much more useful for classification than middle layers.

We further visualize the proportion of unpruned blocks after training schedule on different network architecture and pruning ratio in Figure 7, where for all structures we plot the ratio of unpruned blocks (in blue) in three stages. It is evident that our pruning method prefers to prune blocks in first two stages especially for shallower structures, which is consistent with the pattern in Figure 4, and thus supports the effectiveness of our method.

\begin{figure}[t]
	\centering
	\includegraphics[width=0.9\columnwidth]{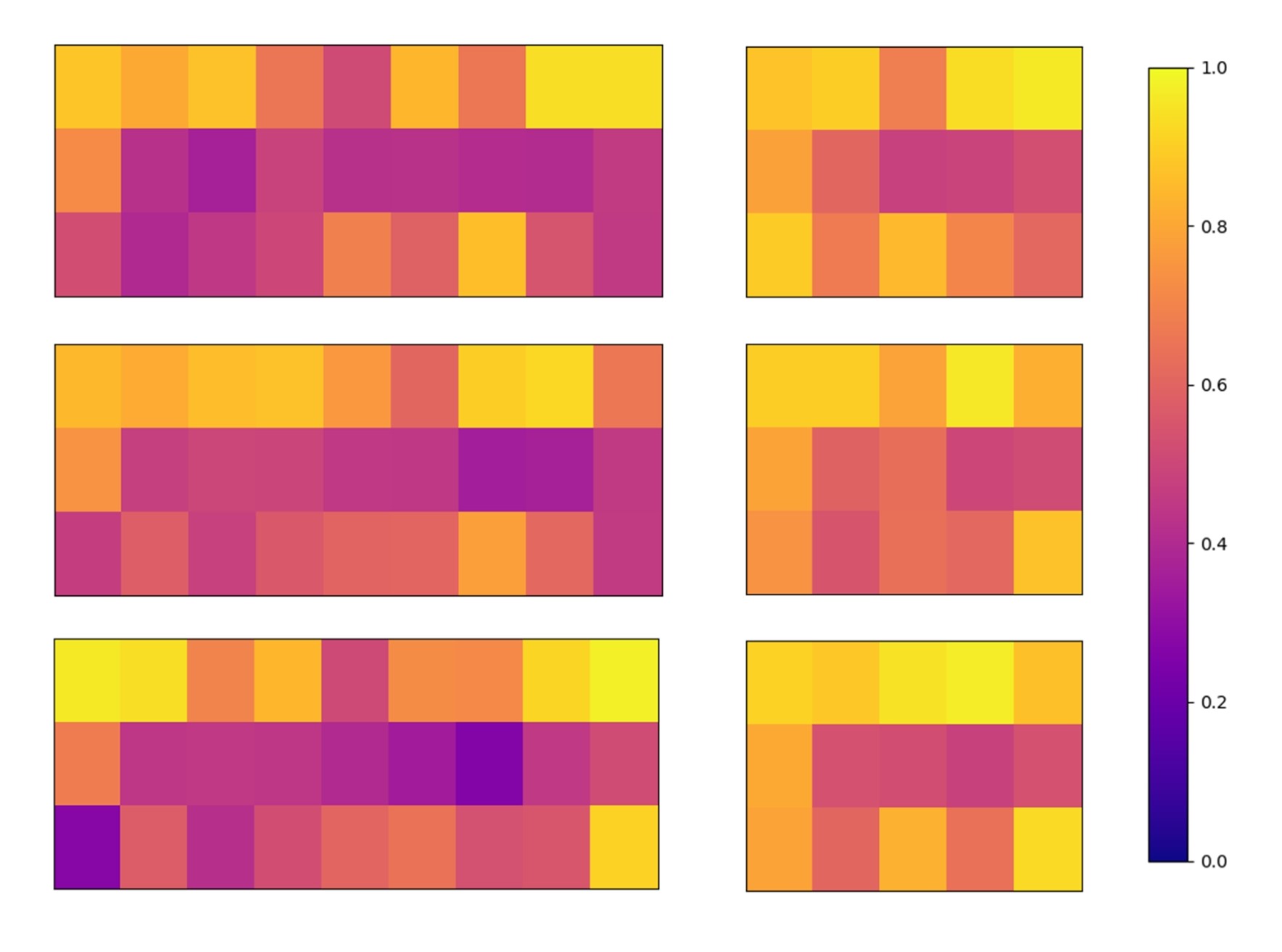} 
	\caption{Visualization of importance marks of three ResNet56 and two ResNet32 models respectively using our methods. Each mesh is divided into three rows, indicating three stages in the architecture, where the lower left corner is the first block and the upper right corner is the last block. Lighter blocks means the mark of this block is larger.}
	\label{fig6}
\end{figure}

\begin{figure}[t]
	\centering
	\includegraphics[width=0.99\columnwidth]{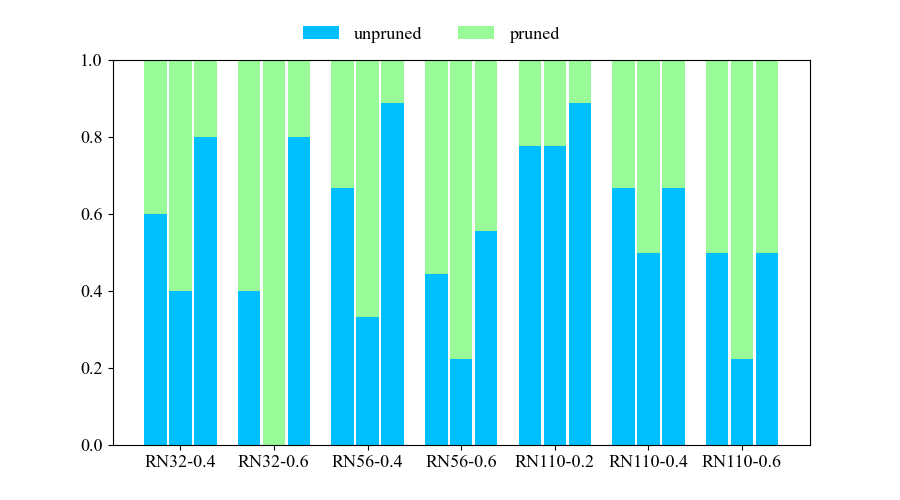} 
	\caption{Proportion of unpruned blocks in different stages for network structures used in  experiments in Table 1. RN stands for ResNet.}
	\label{fig7}
\end{figure}



\subsection{Ablation Study}

In order to validate the effectiveness of different modules of ABP methods, we implemented ablation study on training schedule and knowledge distillation. We successively removed the self-distillation and the whole iterative training schedule in stage 2, which is marked w/o KD and w/o Stage2 respectively in Table 2. Results show that directly pruning according to the importance mark generated through stage 1 training suffers from accuracy drop (94.10\% vs 94.33\% in ResNet110), but still performs better than baseline model with a small pruning rate (0.4). Compared to models trained without self-distillation schema, full ABP training schedule does better in performance recovery after pruning in a large pruning ratio (92.46\% vs 93.28\% in ResNet56-0.6). Note that self-distillation schema is incorporated into the entire training process rather than acting as a training trick, which is similar as GAL\cite{DBLP:conf/cvpr/LinJYZCYHD19}.

\begin{table}[t]
\centering
\begin{tabular}{@{}l|rr|rr@{}}
\toprule
Architecture  & \multicolumn{2}{c|}{ResNet56} & \multicolumn{2}{c}{ResNet110} \\ \midrule
Pruning ratio & 0.4         & 0.6         & 0.4         & 0.6         \\ \midrule
Baseline      & 93.36       & 93.36       & 93.79       & 93.79       \\
w/o Stage2    & 93.47        & 92.46       & 94.10        & 93.46            \\
w/o KD        & 93.56        & 92.59       & 94.33       &  93.69      \\
Full ABP      & 93.70        & 93.28       & 94.41       & 93.85       \\ \bottomrule
\end{tabular}
\caption{Ablation study on different settings used in ABP methods.}
\end{table}




\section{Conclusion}
In this paper we proposed an automatic block-wise pruning schedule for CNNs using gating modules. The core idea of our method is to assign importance marks for all blocks using gating modules, and then prune blocks with lower marks iteratively. By implementing our proposed training schedule, we show that we can cut down over 60\% computational cost with insignificant loss in accuracy for various ResNet models. We further demonstrate how our pruned models can improve the performance of other pruning methods greatly by providing pretrained models, where the performance of pruned model is even better than original model with 93.7\% of FLOPs pruned.

\bibliographystyle{named}
\bibliography{ijcai22}


\end{document}


\maketitle

\begin{table}[ht]
    \centering
\begin{tabular}{crrr}
\toprule
Method              & Accuracy & Acc. ↓ & FLOPs ↓ \\ \midrule
RN32                & 68.81    &        &         \\
ABP-0.4             & 65.01    & 3.80    & 41.11\% \\
ABP-0.6             & 61.9     & 6.91   & 61.67\% \\
ABP(40\%)-SFP(30\%) & 70.1     & -1.29  & 65.25\% \\
ABP(40\%)-SFP(50\%) & 69.51    & -0.7   & 78.27\% \\
ABP(60\%)-SFP(70\%) & 65.86    & 2.95   & 92.82\% \\ \midrule
RN56                & 69.61    &        &         \\
ABP-0.4             & 68.56    & 1.05   & 37.60\% \\
ABP-0.6             & 66.98    & 2.63   & 60.16\% \\
ABP(40\%)-SFP(30\%) & 71.32    & -1.71  & 63.03\% \\
ABP(60\%)-SFP(30\%) & 70.68    & -1.07  & 74.22\% \\
ABP(40\%)-SFP(50\%) & 70.75    & -1.14  & 76.79\% \\
ABP(60\%)-SFP(50\%) & 70.72    & -1.11  & 83.84\% \\
ABP(40\%)-SFP(70\%) & 70.13    & -0.52  & 87.99\% \\
ABP(60\%)-SFP(70\%) & 69.88    & -0.27  & 91.66\% \\
ABP(40\%)-HRank     & 69.62    & -0.01  & 61.39\% \\
ABP(60\%)-HRank     & 69.01    & 0.60    & 73.02\% \\\midrule
RN110               & 71.57    &        &         \\
ABP-0.4             & 71.49    & 0.08   & 39.18\% \\
ABP-0.6             & 69.19    & 2.38   & 59.71\% \\
ABP(40\%)-SFP(30\%) & 73.87    & -2.30   & 62.78\% \\
ABP(60\%)-SFP(30\%) & 72.25    & -0.68  & 76.11\% \\
ABP(40\%)-SFP(50\%) & 73.42    & -1.85  & 76.59\% \\
ABP(60\%)-SFP(50\%) & 71.85    & -0.28  & 84.99\% \\
ABP(40\%)-SFP(70\%) & 72.39    & -0.82  & 87.86\% \\
ABP(60\%)-SFP(70\%) & 70.92    & 0.65   & 92.22\% \\
ABP(40\%)-HRank     & 71.55    & 0.02   & 61.25\% \\
ABP(60\%)-HRank     & 70.31    & 1.26   & 74.95\% \\ \bottomrule
\end{tabular}
\caption{Pruning results on CIFAR100.}
\label{table1}
\end{table}

\section{Incorporating with SOTA methods}

Since ABP is a block-wise pruning method, redundancy within blocks can still be removed for efficient usage. In this section we show that ABP can provide a well designed pre-trained model for other channel pruning methods which leads to much better performance. For fair comparison, we use the baseline performance reported in corresponding papers. Note that the calculated FLOPs below is different from FLOPs reported in previews sections because we use the same methods as the original papers do to calculate computational cost for comparison efficiency.

\begin{algorithm}[t]
    \SetKwData{Left}{left}\SetKwData{This}{this}\SetKwData{Up}{up}
    \SetKwFunction{Union}{Union}\SetKwFunction{FindCompress}{FindCompress}
    \SetKwInOut{Input}{Input}\SetKwInOut{Output}{Output}
  
    \Input{Dataset, $\left \{ (x^1, x^2,...x^M),(y^1, y^2,...y^M) \right \}$;\\     
        Epoch numbers in each stage $E_1$,$E_2$,$E_3$;\\
        Randomly initialized model with and \\
        blocks ${B}=\left \{ {B}_{1},{B}_{2},...,{B}_{N} \right \}$ \\
        gating modules  ${G}=\left \{ {G}_{1},{G}_{2},...,{G}_{N} \right \}$;\\
        Learning rate $\alpha$; Pruning rate $\gamma$;\\
        Block number to prune in each iteration $k$;\\
        Masks initialized to 0 for pruning $\mathcal{S}$;}
    \Output{Learned parameters $W$;\\
        Learned noise distribution $Q(\theta)$ }
    \BlankLine
  \emph{\textbf{Stage \uppercase\expandafter{\romannumeral1}.}}
  
  \While{$epoch < E_1$}{
      Train the model by minimizing $\mathcal{L}_{stage1}$ (6);
  }
  
  \emph{\textbf{Stage \uppercase\expandafter{\romannumeral2}.}}
  
  \While{\textit{block number is greater than }$(1-\gamma) N$}{
      Calculate final mark $\mathcal{M}_{l}$ for each block using (7);\\
      Remove $k$ unpruned blocks with lowest marks by setting corresponding $\mathcal{S}_l$ to 0;
      
      \While{$epoch < E_2$}{
      Train the model by minimizing $\mathcal{L}_{stage2}$ (9);
      }
  }
  
  \emph{\textbf{Stage \uppercase\expandafter{\romannumeral3}.}}
  
  Fix unpruned blocks by setting corresponding $\mathcal{S}_l$ to 1;
  
  \While{$epoch < E_3$}{
      Train the model by minimizing $\mathcal{L}_{stage3}$ (6);
  }
  
    \caption{Training schedule for ABP.}\label{alg.1}
\end{algorithm}

\begin{table*}[h]
    \centering
\begin{tabular}{c|c|ccccc}
\hline
& Architecture  & Method              & Accuracy(\%) & Acc. drop(\%) & FLOPs   & FLOPs drop(\%) \\ \hline
 \multirow{18}{*}{ABP-SFP}& \multirow{6}{*}{ResNet32}  & Baseline  & 92.63             & -     & 68.86M  & -       \\
  &                          & SFP(10\%) & \textbf{93.22} & -0.59    & 58.58M  & 15.2  \\
   &                         & SFP(30\%) & 90.08         & 2.55       & 40.27M  & 41.53  \\
&        & ABP(40\%)-SFP(30\%)           & 91.72         & 0.91      & 21.12M  & 69.33  \\
&        & ABP(40\%)-SFP(50\%)           & 91.46         & 1.17      & 13.20M  & 80.84   \\
&        & ABP(40\%)-SFP(70\%)           & 90.68         & 1.99      & 6.79M   &\textbf{90.68}          \\ \cline{2-7} 
&\multirow{6}{*}{ResNet56}          & Baseline            & 93.59        & -             & 125.49M & -   \\
&          & SFP(10\%)           & 93.89        & -0.30         & 106.99M & 14.74           \\
&  & SFP(40\%)           & 93.35        & 0.24          & 59.43M  & 52.64           \\
&          & ABP(40\%)-SFP(10\%) & \textbf{94.05}        & -0.46         & 58.79M  & 53.15          \\
&          & ABP(40\%)-SFP(50\%) & 93.77        & -0.18         & 25.58M  & 79.61        \\
&          & ABP(60\%)-SFP(70\%) & 92.49        & 1.10          & 7.71M   & \textbf{93.86}          \\ \cline{2-7} 
&\multirow{6}{*}{ResNet110}          & Baseline            & 93.68        & -             & 252.89M & - \\
&          & SFP(20\%)           & 93.93        & -0.25         & 181.55M & 28.21           \\
& & SFP(30\%)           & 93.86        & -0.18         & 149.77M & 40.78           \\
&          & ABP(20\%)-SFP(30\%) & \textbf{94.02}        & -0.34         & 116.58M & 53.90          \\
&          & ABP(20\%)-SFP(50\%) & 93.96        & -0.28         & 73.36M  & 70.99          \\
&          & ABP(60\%)-SFP(70\%) & 93.88        & -0.20         & 15.99M  & \textbf{93.68}          \\ \hline

 \multirow{11}{*}{ABP-HRank} & \multirow{5}{*}{ResNet56}& baseline  & 93.26    &  -  & 125.49M & -              \\
&& HRank     & 93.17    & 0.09  & 62.72M  & 50.02             \\
&& ABP-HRank & 92.35    & 0.91      & 42.70M  & 65.97          \\
&& HRank     & 90.72    & 2.54      & 32.52M  & 74.08           \\
&& ABP-HRank & 91.13    & 2.13      & 26.16M  & 79.15          \\ \cline{2-7} 

 & \multirow{6}{*}{ResNet110}& baseline & 93.50    & -    & 252.89M & -              \\
&& HRank     & 94.23    & -0.73      & 148.70M & 41.19           \\
&& ABP-HRank & 93.53    & -0.03      & 121.74M & 51.86          \\
&& HRank     & 93.36    & 0.14      & 105.70M & 58.20           \\
&& HRank     & 92.56    & 0.94      & 79.30M  & 68.64           \\
&& ABP-HRank & 93.12    & 0.38      & 77.95M  & 69.18          \\ \hline

\end{tabular}
\caption{Results when incorporating our method with SFP \cite{ijcai2018-309} and HRank \cite{DBLP:conf/cvpr/LinJWZZ0020} methods. Percentage inside parentheses after ABP means the pruning ratio in our methods, and percentage after SFP stands for the proportion of pruned channels. Minus values in column "Acc. drop" means increase in accuracy after pruning.}
\label{table3}
\end{table*}

We first implement SFP \cite{ijcai2018-309} method on our pruned model, whose results are shown in Table 2. For each architecture we list the best results reported in original paper, followed with representative results when combining with our ABP method. It is indicated that SFP achieves best accuracy when pruning a small amount of channels on shallow architecture (ResNet32), but suffers accuracy descent with high pruning rate, which is greatly relieved in ABP-SFP method. In experiments for ResNet56, we have achieved even better performance than original SFP method with half of computational cost reduced. Surprisingly on ResNet110 architecture we can achieve even higher accuracy than baseline model after cutting off 93.68\% of FLOPs.

We also apply HRank \cite{DBLP:conf/cvpr/LinJWZZ0020} method on pruned models, and the results are listed in the lower part of Table 2. For both of the architectures, we directly use their pruning rate provided for ResNet110 where all pruning rates are set to 0.4 except for the first convolution layer. In experiments on ResNet56, although it fails to get better performance than original HRank method with small pruning ratio by using out models, the accuracy with high pruning rate can remain comparative higher. The same results also appear on ResNet110, where the accuracy drop is insignificant when pruning on our model with a compression rate of 69.18\%.

\begin{table*}[ht]
    \centering
        \begin{tabular}{cccccc}
            \hline
            Model Name    & Accuracy(\%) & Acc. Drop(\%) & FLOPs   & Params & Inference Time \\ \hline
            ResNet32      & 92.36    & -         & 68.86M  & 0.47M  & 5.63ms         \\
            RecurABP 0.4 & 92.31    & 0.05      & 40.55M  & 0.33M  & 3.41ms         \\
            RecurABP 0.6 & 90.92    & 1.44      & 26.39M  & 0.29M  & 2.97ms         \\
            ConvABP 0.4 & 92.69    & -0.38      & 40.55M  & 0.22M  & 3.10ms         \\
            ConvABP 0.6 & 91.99    & 0.32      & 26.39M  & 0.23M  & 2.39ms         \\ \hline
            ResNet 56     & 93.36    & -         & 125.49M  & 0.86M  & 10.12ms        \\
            RecurABP 0.4 & 93.70    & -0.34      & 78.30M  & 0.66M  & 6.36ms        \\
            RecurABP 0.6 & 93.18    & 0.18      & 49.99M  & 0.41M  & 4.06ms         \\
            ConvABP 0.4 & 93.44    & -0.08      & 78.30M  & 0.52M  & 6.00ms        \\
            ConvABP 0.6 & 93.28    & 0.08      & 49.99M  & 0.48M  & 4.17ms         \\ \hline
            ResNet 110    & 93.79    & -         & 252.89M & 1.73M  & 17.75ms        \\
            RecurABP 0.2 & 94.38    & -0.59     & 205.70M  & 1.49M  & 14.69ms        \\
            RecurABP 0.4 & 94.41    & -0.62      & 153.80M  & 1.09M  & 12.36ms        \\
            RecurABP 0.6 & 93.85    & -0.06     & 101.90M  & 0.76M  & 7.54ms         \\
            ConvABP 0.4 & 93.97    & -0.18      & 153.80M  & 0.95M  & 12.06ms        \\
            ConvABP 0.6 & 92.94    & 0.85      & 101.90M  & 0.61M  & 7.56ms         \\ \hline
        \end{tabular}
\caption{Performance of ABP methods with different pruning ratio and gating strategy(RecurGate and ConvGate).}
\label{table2}
\end{table*}

\section{Result on CIFAR100}
We conducted experiments on CIFAR100 dataset, whose content is the same with CIFAR10 but is divided into 100 classes, which makes it much more difficult for classification. We list the results when using our methods alone and when incorporating with other SOTA pruning methods (SFP and HRank) in Table 1. Results show that ABP models would have significant accuracy drop on small architectures (ResNet32 and ResNet56), but would promote a lot when using them as pre-trained models on SFP methods. For example, such combination can achieve 0.7\% accuracy increase with 78.72\% FLOPs pruned on ResNet32, and on ResNet56, the pruned model can even outperform baseline by 0.27\% with 91.66\% FLOPs off. Moreover, in all three architectures, ABP-SFP can achieve better classification accuracy than baseline except for one with largest pruning ratio. As for HRank, they failed to obtain performance as good as SFP though, they can still cut off over 60\% computational cost with minor accuracy loss. These experiments can verify that ABP can effectively provide prior knowledge for other methods for further pruning.

\section{Pruning ratio and gating types}
The number of blocks to be pruned is decided by pruning rate, which is taken as a hyper-parameter in our work. We investigate the influence brought by pruning ratio by implementing models in various depth with different pruning ratios as shown in Table 2. The result shows that about 40\% of the computational cost would be cut down with a pruning ratio of 0.4 with a minor decrease in accuracy. When the ratio is set to 0.6, the computational cost is reduced to about 1/3 that of baseline, but the accuracy also drops drastically for ResNet32. As for larger models (ResNet56 and ResNet110), the accuracy can even increase with a small pruning ratio(0.2 and 0.4), while pruning 60\% of the blocks would bring less insignificant accuracy drop compared with small models.

We then visualize the accuracy change along with pruning ratio on ResNet56 and ResNet110 architectures to study the effect of pruning ratio on model performance in detail. As shown in Figure 1, all architectures can maintain high accuracy (\textgreater94\%) when using a pruning ratio less than 0.3, and their performance drops slowly before the pruning ratio reaches 0.6. However, when pruning ratio is larger (\textgreater0.6), the accuracy decays rapidly along with the increase of pruning ratio. We can also derive some differences between ABP methods using ConvGate and RecurGate (marked ConvABP and RecurABP) that the accuracy of ConvABP appears to be more severely than RecurABP in total, and RecurABP can thus achieves better performance when pruning ratio is large (\textgreater0.6).

\begin{figure}[t]
	\centering
	\includegraphics[width=0.99\columnwidth]{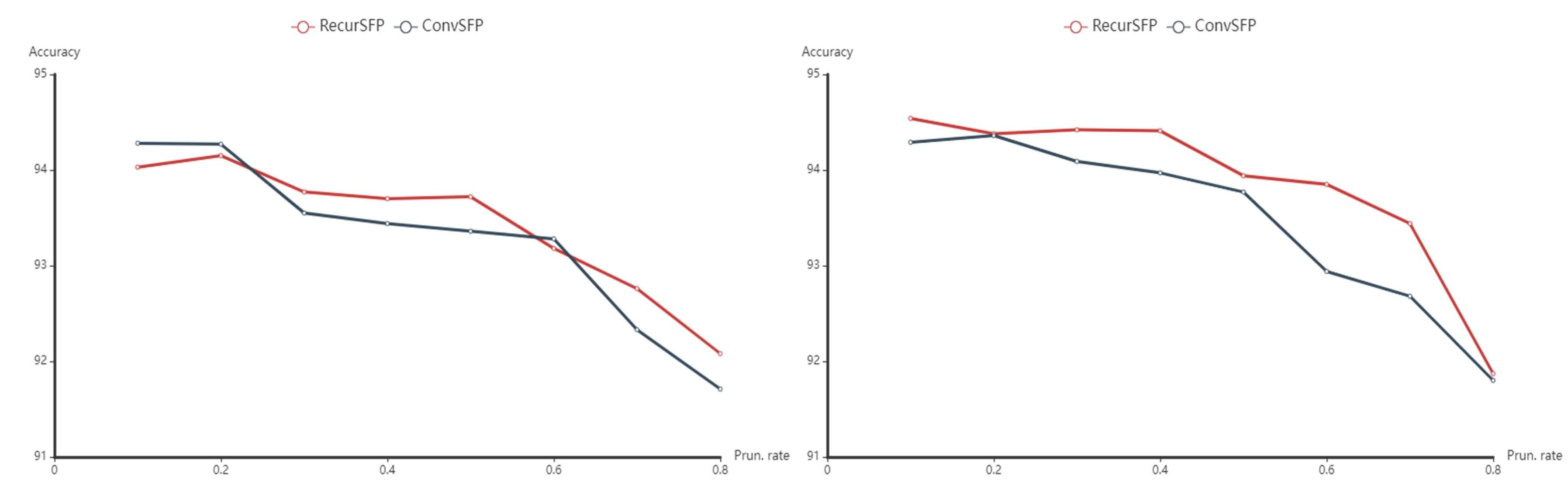} 
	\caption{Accuracy when using different pruning ratio on different gating types (RecurGate and ConvGate), where experiments on ResNet56 is on the left side, while that of ResNet110 is on the right side.}
	\label{fig1}
\end{figure}

\bibliographystyle{named}
\bibliography{ijcai22}